\documentclass{article}

\usepackage{microtype}
\usepackage{graphicx}
\usepackage{subfigure}
\usepackage{booktabs} 
\usepackage{physics}
\usepackage{bbm}

\usepackage[utf8]{inputenc} 
\usepackage[T1]{fontenc}    
\usepackage{hyperref}       
\usepackage{url}            
\usepackage{booktabs}       
\usepackage{amsfonts}       
\usepackage{nicefrac}       
\usepackage{microtype}      
\usepackage{amsmath}
\usepackage{amsthm}
\usepackage{graphicx}
\usepackage{xcolor}
\usepackage{calligra}
\usepackage[font=small,labelfont=bf]{caption}
\usepackage{amsmath}
\usepackage[mathscr]{euscript}
 \let\mathscr\relax
\usepackage[scr]{rsfso}
\newcommand{\powerset}{\raisebox{.15\baselineskip}{\Large\ensuremath{\wp}}}

\usepackage{hyperref}
\usepackage{url}

\newsavebox\IBoxA \newsavebox\IBoxB \newlength\IHeight
\newcommand\TwoFig[6]{
  \sbox\IBoxA{\includegraphics[width=0.45\textwidth]{#1}}
  \sbox\IBoxB{\includegraphics[width=0.45\textwidth]{#4}}%
  \ifdim\ht\IBoxA>\ht\IBoxB
    \setlength\IHeight{\ht\IBoxB}%
  \else\setlength\IHeight{\ht\IBoxA}\fi
  \begin{figure}[h]
  \minipage[t]{0.45\textwidth}\centering
  \includegraphics[height=\IHeight]{#1}
  \caption{#2}\label{#3}
  \endminipage\hfill
  \minipage[t]{0.45\textwidth}\centering
  \includegraphics[height=\IHeight]{#4}
  \caption{#5}\label{#6}
  \endminipage 
  \end{figure}%
}

\DeclareMathOperator*{\argmin}{arg\,min}
\DeclareMathOperator*{\argmax}{arg\,max}
\newtheorem{definition}{Definition}

\usepackage{hyperref}

\usepackage[accepted]{icml2021}

\icmltitlerunning{Causal Curiosity}

\begin{document}

\twocolumn[
\icmltitle{Causal Curiosity: RL Agents Discovering Self-supervised Experiments for Causal Representation Learning}



\icmlsetsymbol{equal}{*}

\begin{icmlauthorlist}
\icmlauthor{Sumedh A Sontakke}{to}
\icmlauthor{Arash Mehrjou}{goo}
\icmlauthor{Laurent Itti}{to}
\icmlauthor{Bernhard Sch{\"o}lkopf}{goo}
\end{icmlauthorlist}

\icmlaffiliation{to}{University of Southern California}
\icmlaffiliation{goo}{Max Planck Institute for Intelligent Systems}

\icmlcorrespondingauthor{Sumedh A Sontakke}{ssontakk@usc.edu}

\icmlkeywords{Causality, Curiosity, Self-supervised Learning, Reinforcement Learning}

\vskip 0.3in
]



\printAffiliationsAndNotice{}  

\begin{abstract}
Animals exhibit an innate ability to learn regularities of the world through interaction. By performing experiments in their environment, they are able to discern the causal factors of variation and infer how they affect the world's dynamics. Inspired by this, we attempt to equip reinforcement learning agents with the ability to perform experiments that facilitate a categorization of the rolled-out trajectories, and to subsequently infer the causal factors of the environment in a hierarchical manner. We introduce {\em causal curiosity}, a novel intrinsic reward, and show that it allows our agents to learn optimal sequences of actions and discover causal factors in the dynamics of the environment. The learned behavior allows the agents to infer a binary quantized representation for the ground-truth causal factors in every environment. Additionally, we find that these experimental behaviors are semantically meaningful (e.g., our agents learn to lift blocks to categorize them by weight), and are learnt in a self-supervised manner with approximately 2.5 times less data than conventional supervised planners. We show that these behaviors can be re-purposed and fine-tuned (e.g., from lifting to pushing or other downstream tasks). Finally, we show that the knowledge of causal factor representations aids zero-shot learning for more complex tasks. Visit \href{https://sites.google.com/usc.edu/causal-curiosity/home}{here} for website.
\end{abstract}

\vspace{-4.5mm}
\section{Introduction}
\label{sec:intro}
\vspace{-1.5mm}

Discovering causation in environments an agent might encounter remains an open and challenging problem for reinforcement learning \cite{bengio2013representation,Schoelkopf15}. In physical systems, causal factors such as gravity or friction affect the outcome of behaviors an agent might perform. Thus, there has been recent interest in attempting to train agents to be robust or invariant against varying values of such causal factors, allowing them to learn modular behaviors that are useful across tasks.
Most model-based approaches take the form of Bayes Adaptive Markov Decision Processes (BAMDPs) \cite{zintgraf2019varibad} or Hidden Parameter MDPs (Hi-Param MDPs) \cite{doshi2016hidden, yao2018direct, killian2017robust, perez2020generalized}
which condition the transition and/or reward function of each environment on hidden parameters.

Formally, let $\vb{s} \in \mathcal{S}$, $\vb{a} \in \mathcal{A}$, $\vb{r} \in \mathcal{R}$, $\vb{h} \in \mathcal{H}$ where $\mathcal{S}$, $\mathcal{A}$, $\mathcal{R}$, and $\mathcal{H}$ are the set of states, actions, rewards and 
admissible causal factors, respectively. In the physical world, examples of the parameter $\vb{h}_j\in\mathcal{H}$ might include gravity, coefficients of friction, masses and sizes of objects.
Hi-Param MDP or BAMDP approaches treat each $\vb{h}_j\in \mathcal{H}$ as a latent variable for which an embedding is learnt during training (often using variational methods \cite{kingma2014semi, ilse2019diva}). Let $\vb{a}_{0:T}$ be a sequence of actions taken by an agent to maximize an external reward resulting in a state trajectory $\vb{s}_{0:T}$. The above approaches define a probability distribution over the entire observable sequence (i.e., rewards, states, actions) as $p(\vb{r}_{0:T}, \vb{s}_{0:T} , \vb{a}_{0:T-1})$ which factorizes as
\begin{equation*}
\prod_{t=1}^{T-1} p(\vb{r}_{t+1}|\vb{s}_t, \vb{a}_t, \vb{s}_{t+1}, \vb{z})p(\vb{s}_{t+1}|\vb{s}_t, \vb{a}_t, \vb{z})p(\vb{a}_t|\vb{s}_t, \vb{z})
\end{equation*}
conditioned on the latent variable $\vb{z}$, a representation for the unobserved causal factors. At test time, in a new environment, the agent infers $\vb{z}$ by observing the trajectories produced by its initial actions issued by the latent conditioned policy obtained during training.

In practice, discovering causal factors in a physical environment is prone to various challenges that are due to the disjointed nature of the influence of these factors on the produced trajectories.
More specifically, at each time step, the transition function is affected by a subset of global causal factors. This subset is implicitly defined on the basis of the current state and the action taken. For example, if a body in an environment loses contact with the ground, the coefficient of friction between the body and the ground no longer affects the outcome of any action that is taken. Likewise, the outcome of an upward force applied by the agent to a body on the ground is unaffected by the friction coefficient. 

Without knowledge of how independent causal mechanisms affect the outcome of a particular action in a given state in an environment, it becomes impossible for the agent to conclude where an encountered variation came from. Unsurprisingly, Hi-Param and BAMDP approaches fail to learn a disentangled embedding of the causal factors, making their behaviors uninterpretable. For example, if, in an environment, a body remains stationary under a particular force, the Hi-Param or BAMDP agent may apply a higher force to achieve its goal of perhaps moving the body, but will be unable to conclude whether the "un-movability" was caused by a high friction coefficient, or high mass.  Additionally, these approaches require substantial reward engineering, making it difficult to apply them outside the simulated environments they are tested in.

\begin{figure*}[t]
    \centering
    \includegraphics[scale=0.4]{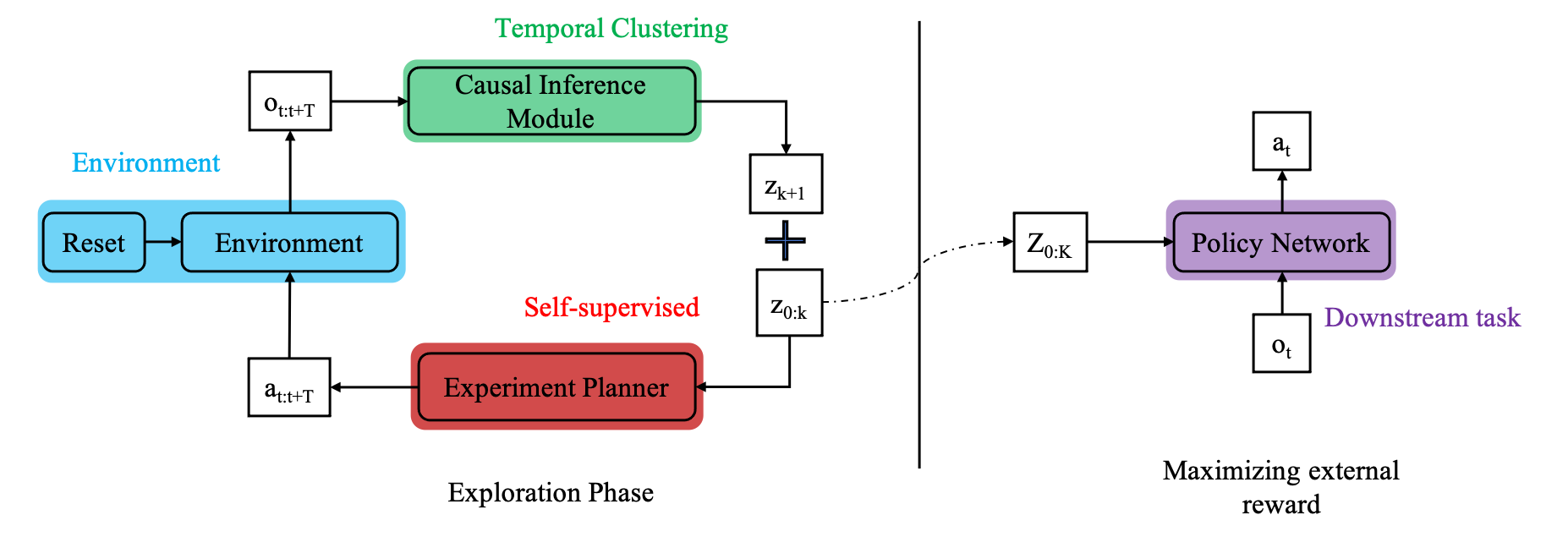}
    \caption{Overview of Inference. The exploration loop produces a series of $K$ experiments allowing the agent to infer the representations for $K$ causal factors. After exploration, the agent utilizes the acquired knowledge for downstream tasks. The details for the inference procedure are provided in Supplementary Material Algorithm \ref{alg:inference}.}
    \label{fig:InferenceLoop}
\end{figure*}

Our goal is, instead of focusing on maximizing reward for a particular task, to allow agents to discover causal processes through exploratory interaction. During training, our agents  discover self-supervised experimental behaviors which they apply to a set of training environments. These behaviors allow them to learn about the various causal mechanisms that govern the transitions in each environment. During inference in a novel environment, they perform these discovered behaviors sequentially and use the outcome of each behavior to infer the embedding for a single causal factor (Figure \ref{fig:InferenceLoop}), allowing us to recover a disentangled embedding describing the causal factors of an environment.


The main challenge while learning a disentangled representation for the causal factors of the world is that several causal factors may affect the outcome of behaviors in each environment. For example, when pushing a body on the ground, the outcome, i.e., whether the body moves, or how far the body is pushed, depends on several factors, e.g., mass, shape and size, frictional coefficients, etc. However, if, instead of pushing on the ground, the agent executes a perfect grasp-and-lift behavior, only mass will affect whether the body is lifted off the ground or not.

Thus, it is clear that not all experimental behaviors are created equal and that the outcomes of some behaviors are caused by fewer causal factors than others. Our agents learn these behaviors without supervision using \textit{causal curiosity}, an intrinsic reward. The outcome of a single such experimental behavior is then used to infer a binary quantized embedding describing the single isolated causal factor. While causal factors of variation in a physical world are easily identifiable to humans, a concrete definition is required to formalize our proposed method. 

\begin{definition}[Causal factors]
  \label{def:causal_factors}
  Consider the POMDP ($\mathcal{O}$, $\mathcal{S}$, $\mathcal{A}$, $\phi$, $\theta$, r) with observation space $\mathcal{O}$, state space $\mathcal{S}$, action space $\mathcal{A}$, the transition function $\phi$, emission function $\theta$, and the reward function $r$. Let  $\vb{o}_{0:T}\in \mathcal{O}^T$ denote a trajectory of observations of length $T$. Let $d(\cdot, \cdot):\mathcal{O}^T \times \mathcal{O}^T\to \mathbb{R}_+$ be a distance function defined on the space of trajectories of length $T$. The set $H=\{\vb{h}_0,\vb{h}_1, \ldots, \vb{h}_{K-1}\}$ is called a set of $\epsilon-$causal factors if for every $\vb{h}_j\in H$, there exists a unique sequence of actions $\vb{a}_{0:T}$ that clusters the observation trajectories into $m$ disjoint sets $C_{1:m}$ such that $\forall C_a, C_b$, a minimum separation distance of $\epsilon$ is ensured:
\begin{equation}
  \min \{d(\vb{o}_{0:T}, \vb{o}'_{0:T}): \vb{o}_{0:T}\in C_a, \vb{o}'_{0:T}\in C_b\} > \epsilon
\end{equation}
and that $\vb{h}_j$ is the cause of the obtained trajectory of states i.e. $\forall v \neq v'$,
\begin{equation}
  p(\vb{o}_{0:T}|do(\vb{h}_j = v), \vb{a}_{0:T}) \neq p(\vb{o}_{0:T}|do(\vb{h}_j = v'), \vb{a}_{0:T})
\end{equation}
where $do(\vb{h}_j)$ corresponds to an intervention on the value of the causal factor $\vb{h}_j$.


\end{definition}
According to Def.~\ref{def:causal_factors}, a causal factor $\vb{h}_j$ is a variable in the environment the value of which, when intervened on (i.e., varied) using $do(\vb{h}_j)$ over a set of values, results in trajectories of observations that are divisible into disjoint clusters $C_{1:m}$ under a particular sequence of actions $\vb{a}_{0:T}$. These clusters represent the quantized values of the causal factor. For example, mass, which is a causal factor of a body, under an action sequence of a grasping and lifting motion, may result in 2 clusters, liftable (low mass) and not-liftable (high mass). 

However, such an action sequence is not known in advance. Therefore, discovering a causal factor in the environment boils down to finding a sequence of actions that makes the effect of that factor prominent by producing clustered trajectories for different values of that environmental factor. For simplicity, here we assume binary clusters.
For a gentle introduction to the intuition about this definition, we refer the reader to Appendix \ref{Gentle}. For an introduction to causality and $do(\cdot)$ notation, see \cite{pearl2009causality, spirtes2010introduction, scholkopf2019causality, elwert2013graphical}.


Our contributions of our work are as follows:
\begin{itemize}
    \item\textbf{Causal POMDPs:} We extend Partially Observable Markov Decision Processes (POMDPs) by explicitly modelling the effect of causal factors on observations.   
    \item \textbf{Unsupervised Behavior:} We equip agents with the ability to perform experiments and behave in a semantically meaningful manner in a set of environments in an unsupervised manner. These behaviors can expose or obfuscate specific independent causal mechanisms that occur in the world surrounding the agent, allowing the agent to "experiment" and learn.
    \item \textbf{Disentangled Representation Learning:} We introduce an minimalistic intrinsic reward, \emph{causal curiosity}, which allows our agents to discover these behaviors without human-engineered complex rewards. The outcomes of the experiments are used to learn a disentangled quantized binary representation for the causal factors of the environment, analogous to the human ability to conclude whether objects are light/heavy, big/small etc. 
    \item \textbf{Sample Efficiency:} Through extensive experiments, we conclude that knowledge of the causal factors aids sample efficiency in two ways - first, that the knowledge of the causal factors aids transfer learning across multiple environments; second, the learned experimental behaviors can be re-purposed for downstream tasks.
\end{itemize}

\section{Method}
Consider a set of $N$ environments $\mathcal{E}$ with $\vb{e}^{i} \in \mathcal{E}$ where $\vb{e}^{i}$ denotes the $i^{th}$ environment. Each causal factor $\vb{h}_j \in H$ is itself a random variable which assumes a particular value for every instantiation of an environment. Thus, every environment $\vb{e}^{i}$ is represented with the values assumed by its causal factors $\{\vb{h}_j^{i}, j=0,1,\ldots, K-1\}$. For each environment $\vb{e}^{i}$, $(\vb{z}_{0}^{i},\vb{z}_{1}^{i}... \vb{z}_{K-1}^{i})$ represents the disentangled embedding vector corresponding to the physical causal factors where $\vb{z}_{j}^{i}$ encodes $\vb{h}_j^{i}$. 

\subsection{POMDP Setup}
\subsubsection{Classical POMDPs}
Classical POMDPs ($\mathcal{O}$, $\mathcal{S}$, $\mathcal{A}$, $\phi$, $\theta$, r) consist of an observation space $\mathcal{O}$, state space $\mathcal{S}$, action space $\mathcal{A}$, the transition function $\phi$, emission function $\theta$, and the reward function $r$. An agent in an unobserved state $\vb{s}_t$ takes an action $\vb{a}_t$ and consequently causes a transition in the environment through $\phi(\vb{s}_{t+1}|\vb{s}_t, \vb{a}_t)$. The agent receives an observation $\vb{o}_{t+1} = \theta(\vb{s}_{t+1})$ and a reward $\vb{r}_{t+1} = r(\vb{s}_t, \vb{a}_t)$. 

\subsubsection{Causal POMDPs}
Our work divides the unobserved state $\vb{s} \in \mathcal{S}$ at each timestep into two portions - the \emph{controllable state} $\vb{s}^c$ and the \emph{uncontrollable state} $\vb{s}^u$. The uncontrollable portion of the state $\vb{s}^u$ consists of the causal factors of the environment. 
We assume that these remain constant during the interaction of the agent with a single instance of the environment. For example, the value of the gravitational acceleration does not change for a single environment. For the following discussion, we refer to the uncontrollable state as causal factors as in Def \ref{def:causal_factors} i.e., $\vb{s}^u$~=~$\mathcal{H}$.

The controllable state $\vb{s}^c$ consists of state variables such as positions and orientations of objects, location of end-effectors of the agent etc. 
Thus, by executing particular action sequences the agent can manipulate this portion of the state, which is hence controllable by the agent.

\subsubsection{Transition Probability}
A trajectory of the controllable state is dependent on both the action sequence that the agent executes and a subset of the causal factors. At each time step, only a subset of the causal factors of an environment affect the transition in the environment. This subset is implicitly selected by the employed policy for every state of the trajectory (depicted as a Gated Causal Graph (Figure \ref{fig:gating})). 
For example, the outcome of an upward force applied by the agent to a body on the ground is unaffected by the friction coefficient between the body and the ground.   

Thus the transition function of the controllable state is:
\begin{equation}
\label{eq:transition}
\phi(\vb{s}^c_{t+1}| \vb{s}^c_{t}, f_{sel}(\vb{\mathcal{H}}, \vb{s}^c_{t}, \vb{a}_{t}),\vb{a}_{t})    
\end{equation}
where $f_{sel}$ is the implicit Causal Selector Function which selects the subset of causal factors affecting the transition defined as:
\begin{equation}
\label{eq:selector}
 f_{sel}:\mathcal{H} \times \mathcal{S} \times \mathcal{A} \to \powerset(\mathcal{H})
\end{equation}
where $\powerset(\mathcal{H})$ is power-set of $\mathcal{H}$ and $f_{sel}(\mathcal{H}, \vb{s}^c_{t}, \vb{a}_{t}) \subset \mathcal{H}$ is the set of effective causal factors for the transition $\vb{s}_t \to \vb{s}_{t+1}$ i.e., $\forall v\neq v'$ and $\forall\vb{h}_j\in f_{sel}(\mathcal{H}, \vb{s}^c_{t}, \vb{a}_{t})$: 
\begin{equation}
  \phi(\vb{s}^c_{t+1}|do(\vb{h}_j = v), \vb{s}^c_{t}, \vb{a}_{t}) \neq \phi(\vb{s}^c_{t+1}|do(\vb{h}_j = v'), \vb{s}^c_{t}, \vb{a}_{t})
\end{equation}

where $do(\vb{h}_j)$ corresponds to an external intervention on the factor $\vb{h}_j$ in an environment. 

Intuitively, this means that if an agent takes an action $\vb{a}_t$ in the controllable state $\vb{s}^c_t$, the transition to $\vb{s}^c_{t+1}$ is caused by a subset of the causal factors $f_{sel}(\mathcal{H}, \vb{s}^c_{t}, \vb{a}_{t})$. For example, if a body on the ground (i.e., state $\vb{s}^c_t$) is thrown upwards (i.e., action $\vb{a}_t$), the outcome $\vb{s}_{t+1}$ is caused by the causal factor gravity (i.e., $f_{sel}(\mathcal{H}, \vb{s}^c_{t}, \vb{a}_{t})=\{\text{gravity}\}$), a singleton subset of the global set of causal factors. The $do()$ notation expresses this causation. If an external intervention on a causal factor is performed, e.g., if somehow the value of gravity was changed from $v$ to $v'$, the outcome of throwing the body up from the ground, $\vb{s}_{t+1}$, would be different. 

\subsubsection{Emission Probability}
The agent neither has access to the controllable state, nor to the causal factors of each environment. It receives an observation described by the function:
\begin{equation}
  \vb{o}_{t+1} = \theta(\vb{s}^c_{t}, f_{sel}(\vb{\mathcal{H}}, \vb{s}^c_{t}, \vb{a}_{t}),\vb{a}_{t})
\end{equation}
where $f_{sel}$ is the implicit Causal Selector Function.


\subsection{Training the Experiment Planner}
The agent has access to a set of training environments with multiple causal factors varying simultaneously.  
Our goal is to allow the agent to discover action sequences $\vb{a}_{0:T}$ such that the resultant observation trajectory $\vb{o}^i_{0:T}$ is caused by a single causal factor i.e., $\forall t<T, f_{sel}(\mathcal{H},\vb{s}^c_{t}, \vb{a}_{t})=\text{constant}~\text{and}~|f_{sel}(\mathcal{H},\vb{s}^c_{t}, \vb{a}_{t})|=1$. Consequently, $\vb{o}^i_{0:T}$ can be used to learn a representation $\vb{z}^{i}_{j}$ for the causal factor $f_{sel}(\mathcal{H},\vb{s}^c_{t}, \vb{a}_{t})$ for each environment $\vb{e}^i$.


\begin{figure}[ht]
\begin{center}
\centerline{\includegraphics[width=\columnwidth]{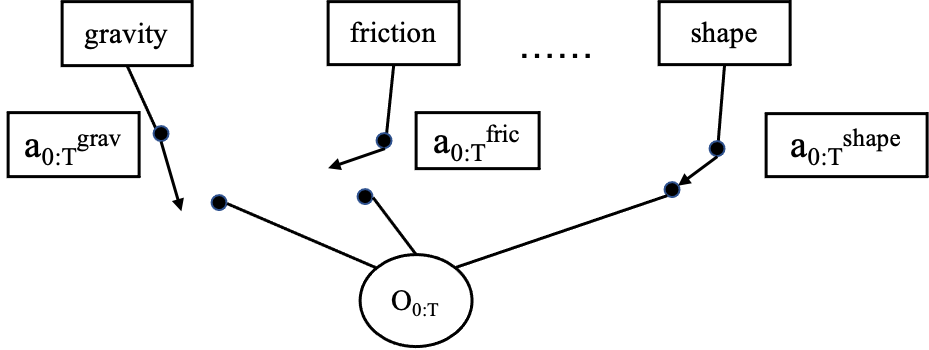}}
\caption{Gated Causal Graph. A subset of the unobserved parent causal variables influence the observed variable $\vb{O}$. The action sequence $\vb{a}_{0:T}$ serves a gating mechanism, allowing or blocking particular edges of the causal graph using the implicit Causal Selector Function (Equation \ref{eq:selector}).}
\label{fig:gating}
\end{center}
\end{figure}
We motivate this from the perspective of algorithmic information theory \citep{JanSch10}. Consider the Gated Directed Acyclic Graph of the observed variable $\vb{O}$ and its causal parents (Figure \ref{fig:gating}). Each causal factor has its own causal mechanism, jointly bringing about $\vb{O}$. A central assumption of our approach is that causal factors are independent, i.e., the Independent Mechanisms Assumption \cite{Schoelkopf2012, parascandolo2018learning, scholkopf2019causality}. The information in $\vb{O}$ is then the sum of information “injected” into it from the multiple causes, since, loosely speaking, for information to cancel, the mechanisms would need to be algorithmically dependent \cite{JanSch10}. Intuitively, the information content in $\vb{O}$ will be greater for a larger number of causal parents in the graph. Interestingly, a similar argument has been made to justify the thermodynamic arrow of time \cite{chaves2014inferring, janzing2016algorithmic}: while a microscopic time evolution is invertible, the assumption of algorithmic independence for initial conditions and forward mechanisms generates an asymmetry. 
To invert time, the backward mechanism would need to depend on the initial state.

Thus we attempt to find an action sequence $\vb{a}_{0:T}$ for which the number of causal parents of the resultant observation $\vb{O}$ is low, i.e., the complexity of $\vb{O}$ is low. One could conceive of this by assuming that the generative model for $\vb{O}$, $\vb{M}$ has low Kolmogorov Complexity. Here, a low capacity bi-modal model is assumed. We utilize Minimum Description Length $L(\cdot)$ (MDL) as a tractable substitute of the Kolmogorov Complexity \cite{rissanen1978modeling, grunwald2004tutorial}).   

Causal curiosity solves the following optimization problem.  
\begin{equation}
\label{eq:new_mdl}
\vb{a}_{0:T}^* = \argmin_{\vb{a}_{0:T}}({L(\vb{M}) + L(\vb{O}|\vb{M})})
\end{equation}
where each observed trajectory $\vb{O}=\vb{O}(\vb{a}_{0:T})$ is a function of the action sequence. As mentioned earlier, the model is fixed in this formulation; hence, the first term $L(\vb{M})$ is constant and not a function of the actions.
The MDL of the trajectories given binary categorization model, $-L(\vb{O}|\vb{M})$, is the inherent reward function that is fed back to the RL agent. We regard this reward function as~\emph{causal curiosity}. See Supplementary Material \ref{TrainingDeets} for  implementation details. 


\begin{algorithm}[tb]
\caption{Recursive Training Scheme}\label{alg:training}
\begin{algorithmic}
\STATE Initialize $j = 0$
\STATE Initialize training environment set $Envs$
\FUNCTION{\textbf{Train}($j$, $\vb{z}^{i}_{0:j}$, $Envs$)}
\IF{$j == K$}
\STATE Return
\ENDIF
\FOR{iteration m to M}
\STATE Sample experimental behavior $\vb{a}_{0:T}$ $\sim$ CEM($\cdot|\vb{z}^{i}_{0:j}$)
\FOR{$i^{th}$ env in $Envs$}
\STATE Apply $\vb{a}_{0:T}$ to env
\STATE Collect $\vb{O}^i = \vb{o}_{0:T}^{i}$
\STATE Reset env
\ENDFOR
\STATE Calculate $-L(\vb{O}|\vb{M})$ given that $\vb{M}$ is bimodal clustering model
\STATE Update CEM($\cdot$) distribution with highest reward trajectories 
\ENDFOR

\STATE Use learnt $q_M(\vb{z}^{i}_{j}|\vb{O},\vb{z}^{i}_{0:j})$ for cluster assignment of each env in $Envs$ i.e. $\vb{z}^{i}_j = q_M(\vb{z}|\vb{O}^{i},\vb{z}^{i}_{0:j})$
\STATE Update $j = j+1$
\STATE \textbf{Train}($j$, $\vb{z}^{i}_{0:j}$, $Envs=\{\vb{e}^{i} : \vb{z}^{i}_{j-1}=0\}$)
\STATE \textbf{Train}($j$, $\vb{z}^{i}_{0:j}$, $Envs = \{\vb{e}^{i} : \vb{z}^{i}_{j-1}=1\}$)

\ENDFUNCTION
\end{algorithmic}
\end{algorithm}

\subsection{Causal Inference Module}
By maximizing the causal curiosity reward it is possible to achieve behaviors which result in trajectories of states only caused by a single hidden parameter. Subsequently, we utilize the outcome of performing these experimental behaviors in each environment to infer a representation for the causal factor isolated by the experiment in question. 

We achieve this through clustering. 
An action sequence $\vb{a}_{0:T}\sim \text{CEM}(\cdot|\vb{z}^{i}_{0:j-1})$ is sampled from the Model Predictive Control Planner \cite{camacho2013model} and applied to each of the training environments. The learnt clustering model $\vb{M}$ is then used to infer a representation for each environment using the collected outcome $\vb{O}^{i}$ obtained by applying $\vb{a}_{0:T}$ to each environment. 
\begin{equation}
\label{eq:representation_learning}
\vb{z}^{i}_j = q_M(\vb{z}|\vb{O}^{i}, \vb{z}^{i}_{0:j-1})
\end{equation}

The learnt representation $\vb{z}$ is the cluster membership obtained from the learnt clustering model $\vb{M}$. It is binary in nature. This corresponds to the quantization of the continuous spectrum of values a causal factor takes in the training set into high and low values.
\subsection{Interventions on beliefs}
\label{sec:Interventions}
Having learnt about the effects of a single causal factor of the environment we wish to learn such experimental behaviors for each of the remaining hidden parameters that may vary in the training environments.
To achieve this, in an ideal setting, the agent would require access to the generative mechanism of the environments it encounters. Ideally, it would hold the values of the causal factor already learnt about a constant i.e. $do(\vb{h}_j = constant)$, and intervene over (vary the value of) another causal factor over a set of values $V$ i.e. $do(\vb{h}_{j'} = v)$ such that $v \in  V$. For example, if a human scientist were to study the effects of a causal factor, say mass of a body, they would hold the values of all other causal factors constant (e.g., interact with cubes of the same size and external texture), and vary only mass to see how it affects the outcome of specific behaviors they apply to each body.  

However, in the real world the agent does not have access to the generative mechanism of the environments it encounters, but merely has the ability to act in them. Thus, it can intervene on the representations of a causal factor of the environment i.e. $do(\vb{z}_i = constant)$. For, example having learnt about gravity, the agent picks all environments it believes have the same gravity, and uses them to learn about a separate causal factor say, friction. 

Thus, to learn about the $j^{th}$ causal factor, the agent proceeds in a tree-like manner, dividing each of the $2^{j-1}$ clusters of training environments into two sub-clusters corresponding to the binary quantized values of the $j^{th}$ causal factor. Each level of this tree corresponds to a single causal factor.
\begin{equation}
\label{eq:interventions}
Envs = \{\vb{e}^{i} : \vb{z}^{i}_{j-1}=k\},  k \in \{0,1\}
\end{equation}
This process continues iteratively (Algorithm \ref{alg:training} and Figure~\ref{fig:hierarchyZ}), where for each cluster of environments, a new experiment learns to split the cluster into 2 sub-clusters depending on the value of a hidden parameter. At level $n$, the agent produces $2^n$ experiments, having already intervened on the binary quantized representations of $n$ causal factors.

\section{Related Work}
Curiosity for robotics is not a new area of research. Pioneered by Schmidhuber in the 1990s, \cite{schmidhuber1991possibility, schmidhuber1991curious, schmidhuber2006developmental,schmidhuber2010formal}, \cite{ngo2012learning}, \cite{pathak2017curiosity} curiosity is described as the motivation behind the behavior of an agent in an environment for which the outcome is unpredictable, i.e., an intrinsic reward that motivates the agent to explore the unseen portions of the state space (and subsequent transitions). 
\cite{doshi2016hidden} define a class Markov Decision Processes where transition probabilities $p(s_{t+1}|s_{t},a_{t};\theta)$ depend on a hidden parameter $\theta$, whose value is not observed, but its effects are felt. \cite{killian2017robust} and \cite{yao2018direct} utilize these Hidden Parameter MDPs (Markov Decision Processes) to enable efficient policy transfer, assuming that transition probabilities across states are a function of hidden parameters. \cite{perez2020generalized} relax this assumption, allowing both transition probabilities and reward functions to be functions of hidden parameters. \cite{zintgraf2019varibad} approach the problem from a Bayes-optimal policy standpoint, defining transition probabilities and reward functions to be dependent on a hidden parameter characteristic of the MDP in consideration. We utilize this setup to define causal factors. 
\newline Substantial attempts have been made at unsupervised disentanglement, most notably, the $\beta$-VAE \cite{higgins2017beta} \cite{burgess2018understanding}, where a combination of factored priors and the information bottleneck force disentangled representations. \cite{kim2018disentangling} enforce explicit factorization of the prior without compromising on the mutual information between the data and latent variables, a shortcoming of the $\beta$-VAE. \cite{chen2018isolating} factor the KL divergence into a more explicit form, highlighting an improved objective function and a classifier-agnostic disentanglement metric. 
\cite{locatello2018challenging} show theoretically that unsupervised disentanglement (in the absence of inductive biases) is impossible and highly unstable, susceptible to random seed values. They follow this up with \cite{locatello2020weakly} where they show, both theoretically and experimentally, that pair-wise images provide sufficient inductive bias to disentangle causal factors of variation. However, these works have been applied to supervised learning problems whereas we attempt to disentangle the effects of hidden variables in dynamical environments, a relatively untouched question.

\section{Experiments}
\begin{figure*}[ht!]
    \centering
    \includegraphics[scale = 0.46]{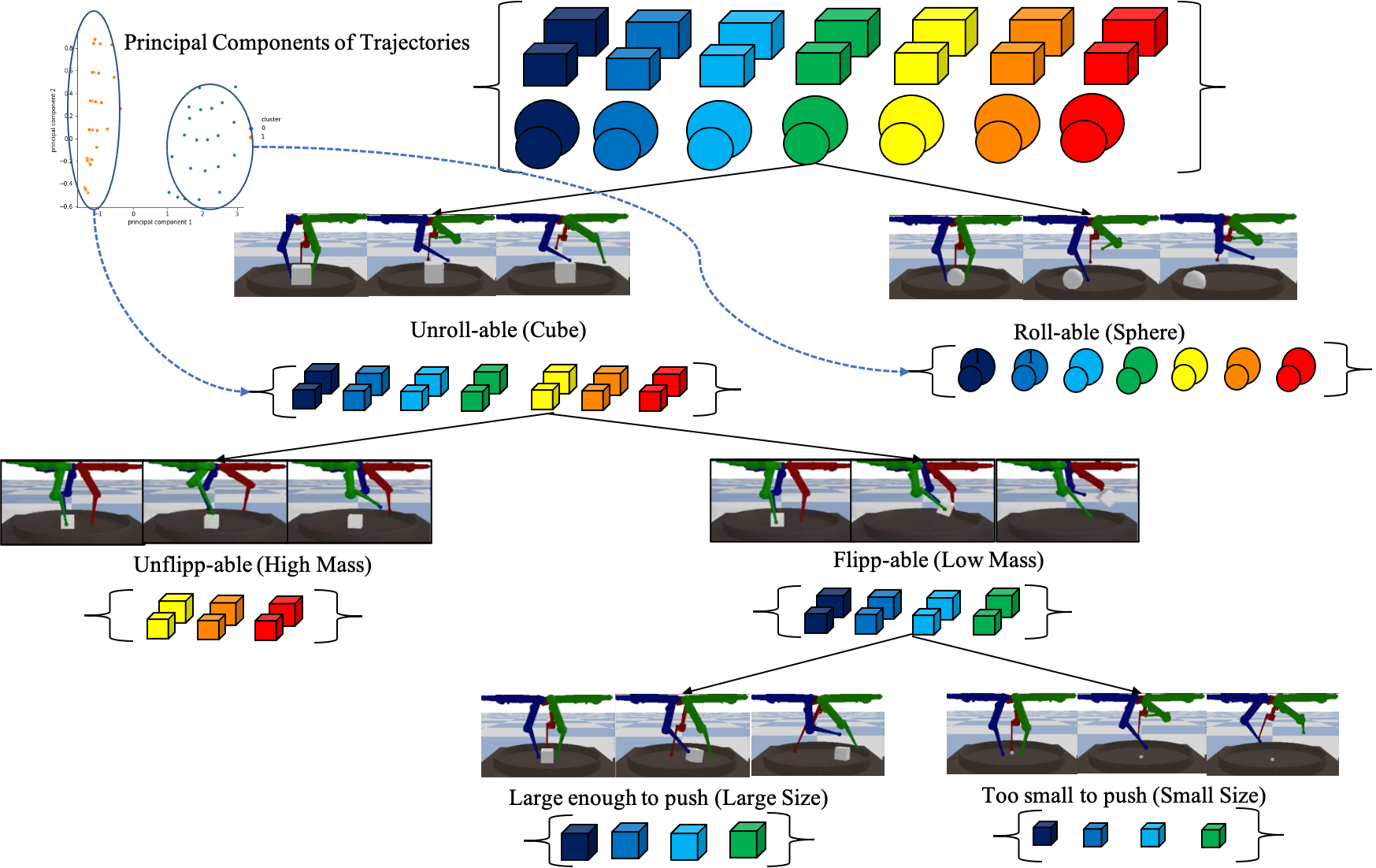}
    \caption{Discovered hierarchical latent space. The agent learns experiments that differentiate the full set of blocks in \texttt{ShapeSizeMass} into hierarchical binary clusters. At each level, the environments are divided into 2 clusters on the basis of the value of a single causal factor. We also show the principal components of the trajectories in the top left. For brevity, the full of extent of the tree is not depicted here. For each level of hierarchy $k$, there are $2^k$ number of clusters.}
    \label{fig:hierarchyZ}
\end{figure*}
Our work has 2 main thrusts - the discovered \emph{experimental behaviors} and the \emph{representations} obtained from the outcome of the behaviors in environments. We visualize these learnt behaviors and verify that they are indeed semantically meaningful and interpretable. We quantify the utility of the learned behaviors by using the behaviors as pre-training for a downstream task. In our experimental setup, we verify that these behaviors are indeed invariant to all other causal factors except one. \newline We visualize the representations obtained using these behaviors and verify that they are indeed the binary quantized representations for each of the ground truth causal factors that we manipulated in our experiments. Finally, we verify that the knowledge of the representation does indeed aid transfer learning and zero-shot generalizability in downstream tasks.

\textbf{Causal World.} We use the Causal World Simulation \cite{ahmed2020causalworld} based on the Pybullet Physics engine to test our approach. The simulator consists of a 3-fingered robot, with 3 joints on each finger. We constrain each environment to consist of a single object that the agent can interact with. The causal factors that we manipulate for each of the objects are size, shape and mass of the blocks. The simulator allows us to capture and track the positions and velocities of each of the movable objects in an environment. 

\textbf{Mujoco Control Suite.} We optimize causal curiosity on 4 articulated agents that try to learn locomotion - Ant, Half-Cheetah, Hopper, and Walker. For each agent type, we train with agent body masses from $0.5\times$ to $1.5\times$ the default.

\subsection{Visualizing Discovered Behaviors}
We would like to analyze whether the discovered experimental behaviors are human interpretable, i.e., \textit{are the experimental behaviors discovered in each of the setups semantically meaningful?} We find that our agents learn to perform several useful behaviors without any supervision. For instance, to differentiate between objects with varying mass, we find that they acquire a perfect grasp-and-lift behavior with an upward force. In other random seed experiments, the agents learn to lift the blocks by using the wall of the environment for support. To differentiate between cubes and spheres, the agent discovers a pushing behavior which gently rolls the spheres along a horizontal direction. Qualitatively, we find that these behaviors are stable and predictable. See videos of discovered behaviors \href{https://drive.google.com/drive/folders/13hZgzW_Tbd5EicLbIxUMQqD7KhfRSW9g?usp=sharing}{\texttt{here}} (website under construction).  

\begin{figure*}[t]
    \centering
    \includegraphics[scale=0.5]{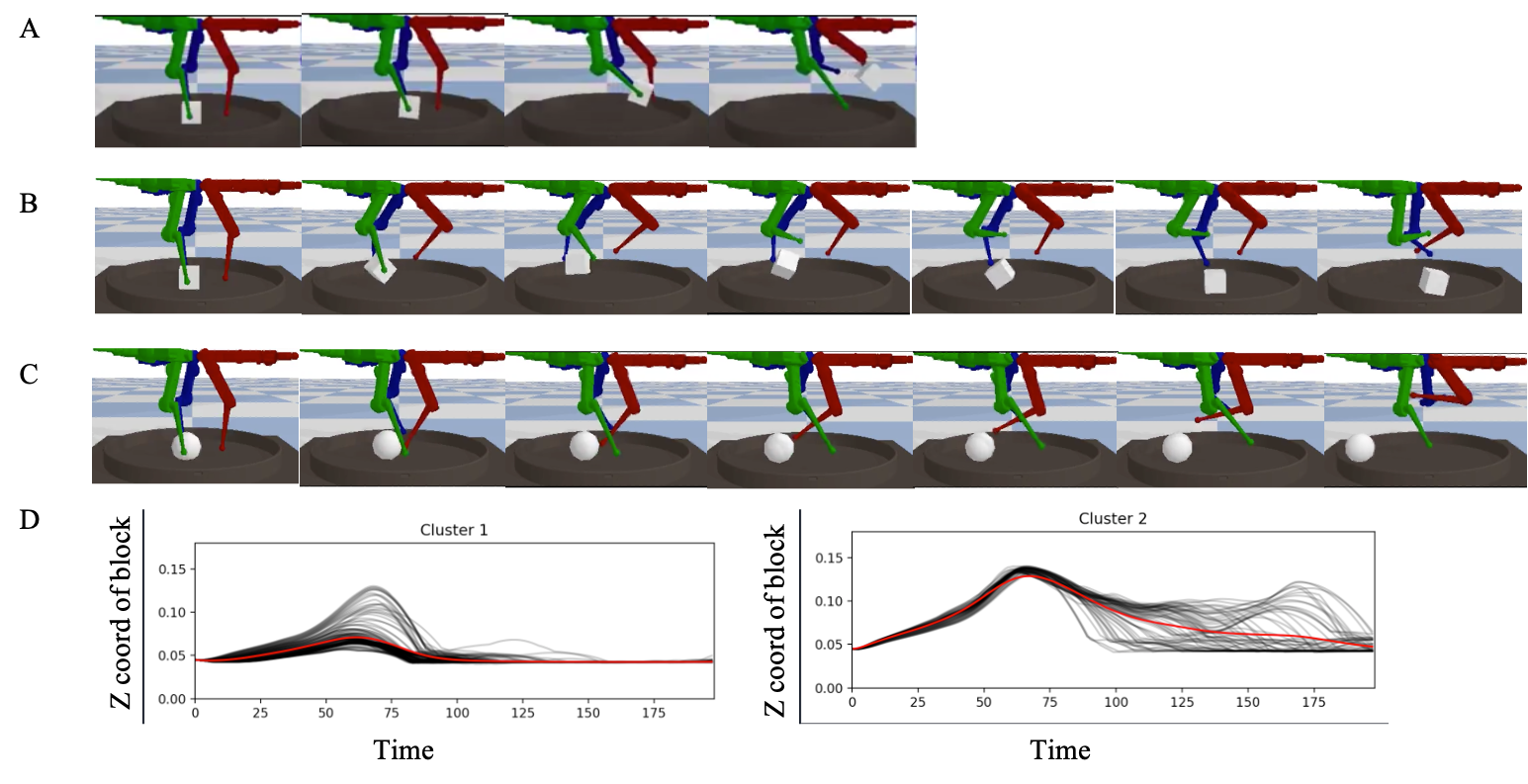}
    \caption{Examples of discovered behaviors. The agent discovers experimental behaviors that allow it to characterize each environmental object in a binary manner, e.g., heavy/light, big small, rollable/not rollable, etc. These behaviors are acquired without any external supervision by maximizing the causal curiosity reward. \textbf{A, B, C} correspond to self-discovered toss, lift-and-spin and roll behaviors respectively. \textbf{D} shows an ablation study, where the agent is only provided the z coordinate of the block in every environment. Each line corresponds to one environment and the z coordinate of the block is plotted with time when the discovered behavior is applied. It learns a lifting behavior, where cluster 1 represents the heavy blocks (z coordinate does not change much) and cluster 2 represents the light blocks (z increases as block is lifted and then falls when dropped and subsequently increases again when it bounces).}
    \label{fig:visBehav}
\end{figure*}

Concurrent with the objective they are trained on, we find that the acquired behaviors impose structure on the outcome when applied to each of the training environments. The outcome of each experimental behavior on the set of training environments results in dividing it into 2 subsets corresponding to the binary quantized values of a single factor, e.g., large or small, while being invariant to the values of other causal factors of the environments. 
We also perform ablation studies where instead of providing the full state vector, we provide only one coordinate (e.g., only x, y or z coordinate of the block). We find that causal curiosity results in behaviors that differentiate the environments based on outcomes along the direction provided. For example, when only the x coordinate was provided, the agent learned to evaluate mass by applying a pushing behavior along the x direction. Similarly, a lifting behavior was obtained when only the z coordinate was supplied to the curiosity module (Figure \ref{fig:visBehav}). 

\begin{figure*}[h!]
    \centering
    \includegraphics[scale=0.5, width=\textwidth]{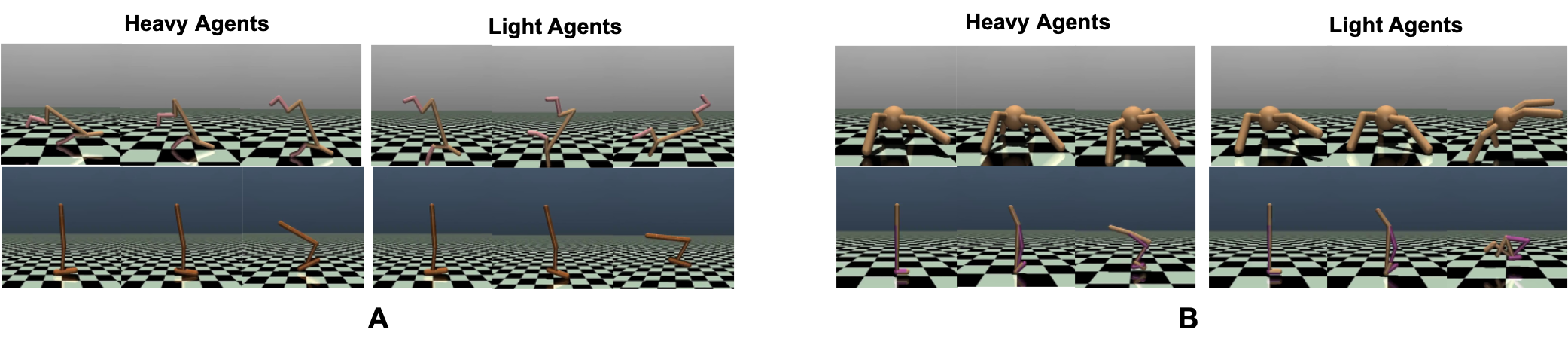}
\caption{\textbf{Mujoco Experiments.} Pane \textbf{A} shows discovered experimental behaviors on Half-Cheetah and Hopper, while Pane \textbf{B} shows discovered experimental behaviors on Ant and Walker. The Half-Cheetah learns to front-flip, the Hopper learns to hop, and the Ant learns to rear up on its hind legs to gauge its mass.\vspace{-1ex}}
    \label{fig:Mujoco}
\end{figure*}

Causal curiosity also yields semantically meaningful behaviors that test out agent mass in Mujoco: the Half-Cheetah learns a front-flip, the Hopper learns to hop to gauge its own mass, in the absence of external rewards (Fig \ref{fig:Mujoco}).

\subsection{Utility of learned behaviors for downstream tasks}
\label{downstream}
\begin{figure*}[h!]
    \centering
    \includegraphics[scale = 0.5]{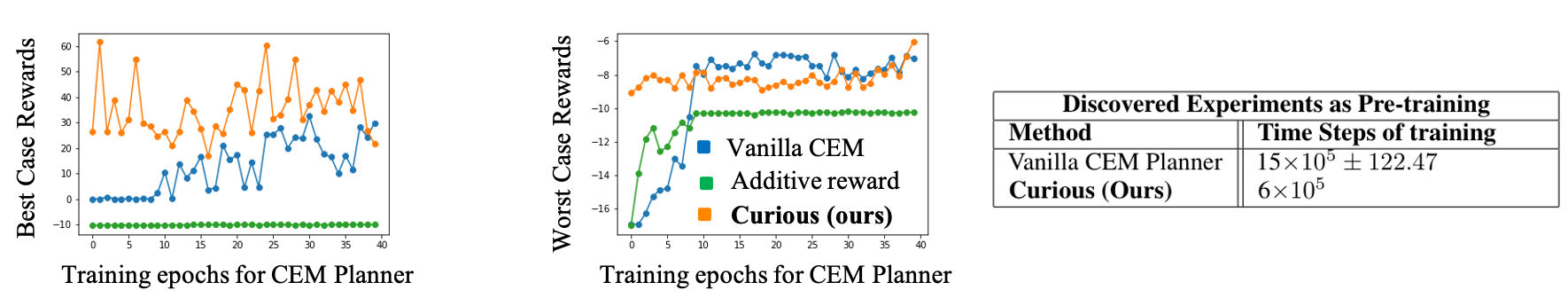}
    \caption{Utility of discovered behaviors.  We find that the behaviors discovered by the agents while optimizing causal curiosity show high zero-shot generalizability and converge to the same performance as conventional planners for downstream tasks. We also analyze the worst case performance and find that the pre-training ensures better performance than random initialization. The table compares the time-steps of training required on an average to acquire a skill with the time steps required to learn a similar behavior using external reward. We find that the unsupervised experimental behaviors are approximately 2.5 times more sample efficient. We also find that maxizing both curiosity and external reward in our experimental setups results in sub-optimal results. }
    \label{fig:ZeroShot}
\end{figure*}

While the behaviors acquired are semantically meaningful, we would like to quantify their utility as pre-training for downstream tasks. We analyze the performance on \texttt{Lifting} where the agent must grasp and lift a block to a predetermined height and \texttt{Travel}, where the agent must impart a velocity to the block along a predetermined direction. We re-train the learnt planner using an external reward for these tasks (\textbf{Curious}). We implement a baseline vanilla Cross Entropy Method optimized Model Predictive Control Planner \cite{de2005tutorial} (\textbf{Vanilla CEM}) trained using the identical reward function and compare the rewards per trajectory during training. We also run a baseline (\textbf{Additive reward}) which explores whether the agent recieves both the causal curiosity reward and the external reward. We find high zero-shot generalizability and quicker convergence as compared to the vanilla CEM (Figure~\ref{fig:ZeroShot}). We also find that additive rewards, achieves suboptimal performance due to competing objectives. For details, we refer the reader to the Supplementary Material \ref{sec:downstreamDetails}. 
\begin{figure*}[h!]
    \centering
    \includegraphics[scale = 0.5]{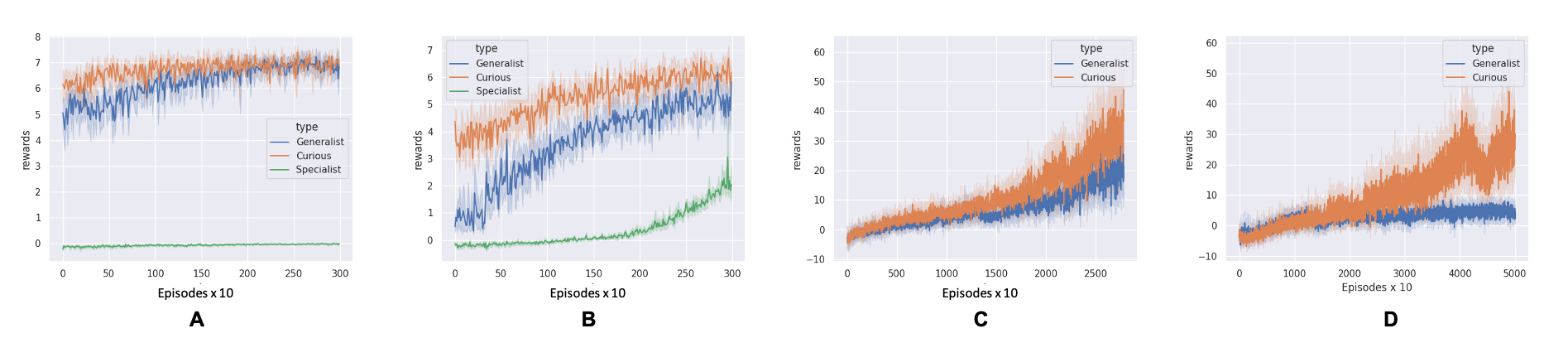}
    \caption{Knowledge of causal factors aids transfer. We find that knowledge of the causal representation allows agents to generalize to unseen environments with high zero-shot performance. We find that as the number of varying causal factors increase, the difference in zero-shot performance of the Causally-curious agent and the Generalist increases, showing that the CC agents are indeed robust to multiple varying causal factors. Panes \textbf{A} and \textbf{B} are represent reward curves for \texttt{TransferMass} and \texttt{TransferSizeMass}. Pane \textbf{C} shows the training rewards curve for the new \texttt{StackingTower} experiment and Pane \textbf{D} shows the reward curves during adaptation to an unseen value of causal factors.}
    \label{fig:outlier_removal}
\end{figure*}

\subsection{Visualization of hierarchical binary latent space}
\label{4.3}
Our agents discover a disentangled latent space such that they are able to isolate the sources of causation of the variability they encounters in their environments. For every environment, they learn a disentangled embedding vector which describes each of the causal factors. 

To show this, we use 3 separate experimental setups - \texttt{Mass}, \texttt{SizeMass} and \texttt{ShapeSizeMass} where each of the causal factors are allowed to vary over a range of discrete values. For details of the setup, we refer the reader to Supplementary Material \ref{subsec:TrainEnv}.

During training, the agent discovers a hierarchical binary latent space (Figure \ref{fig:hierarchyZ}), where each level of hierarchy corresponds to a single causal factor. The binary values at each level of hierarchy correspond to the high/low values of the causal factor in question. To our knowledge, we obtain the first interpretable latent space describing the various causal processes in the environment of an agent. This implies that it learns to quantify each physical attribute of the blocks it encounters in a completely unsupervised manner.

\subsection{Knowledge of causal factors aids transfer}
\label{4.4}
Next, we test whether knowledge of the causal factors does indeed aid transfer and zero-shot generalizability. To this end, we supply the representations obtained by the agent during the experimental behavior phase as input to a policy network in addition to the state of the simulator, and train it for a place-and-orient downstream task (Figure \ref{fig:InferenceLoop}). We define 2 experimental setups - \texttt{TransferMass} and \texttt{TransferSizeMass} where mass and size of the object in each environment is varied. We also test our agent in a separate task, \texttt{StackingTower}, where the agent is provided 2 blocks which it must use to build a stable tower configuration. These blocks vary in mass and the agent must use causal representations to learn to build towers with a heavy base for stability.
In each of the setups, the agent learns about the varying causal mechanisms by optimizing causal curiosity. Subsequently, using the causal representation along with the state for each environment, it is trained to maximize external reward. For details of the setup, please see Supplementary Material \ref{TranDeets}.

After training, the agents are exposed to a set of unseen test environments, where we analyze their zero-shot generalizability. These test environments consist of unseen masses and sizes and their unseen combinations. This corresponds to "Strong Generalization" as defined by \cite{perez2020generalized}. We report results averaged over 10 random seeds. 

For each setup, we train a PPO-optimized Actor-Critic Policy (referred to as \textbf{Causally-curious agent}) with access to the causal representations and an observation vector from the environment i.e., $\vb{a}_t \sim \pi(\cdot | \vb{o}_t, \vb{z}_{0:K})$. Similar to \cite{perez2020generalized}, we implement 2 baselines - the \textbf{Generalist} and the \textbf{Specialist}. The \textbf{Specialist} consists of an agent with identical architecture as \textbf{Causally-curious agent}, but without access to causal representations. It is initialized randomly and is trained only on the test environments, serving as a benchmark for complexity of the test tasks. It performs poorly, indicating that the test tasks are complex. The architecture of the \textbf{Generalist} is identical to the \textbf{Specialist}. Like the \textbf{Specialist}, the \textbf{Generalist} also does not have access to the causal representations, but is trained on the same set of training environments that the Causally-curious agent is trained on. The poor performance of the generalist indicates that the tasks distribution of training and test tasks differs significantly and that memorization of behaviors does not yield good transfer. We find that causally-curious agents significantly outperform the both baselines indicating that indeed, knowledge of the causal representation does aid zero-shot generalizability.

\section{Conclusion}
Our work introduces a causal viewpoint of POMDPs, where unobserved static state variables (i.e., causal factors) affect the transition of dynamic state variables. Causal curiosity rewards experimental behaviors an agent can conduct in an environment that underscore the effects of a subset of such global causal factors while obfuscating the effects of others. Motivated by the popular One-Factor-at-a-Time (OFAT) \cite{fisher1936design,hicks1964fundamental,czitrom1999one}, our agents study the effects causal factors have on the dynamics of an environment through active experimentation and subsequently obtain a disentangled causal representation for causal factors of the environment. We discuss the implication of OFAT in Supplementary Material \ref{sec:ofat}. Finally, we show that knowledge of causal representations does indeed improve sample efficiency in transfer learning. 
\section*{Acknowledgements}
This work was supported by C-BRIC (one of six centers in JUMP, a Semiconductor Research Corporation (SRC) program sponsored by DARPA), the Army Research Office (W911NF2020053), and the Intel and CISCO Corporations. The authors affirm that the views expressed herein are solely their own, and do not represent the views of the United States government or any agency thereof. SAS was partly funded by the Annenberg Fellowship. Many thanks also to Alexander Neitz for sourcing of the CEM planning code. SAS would like to thank Stefan Bauer, Theofanis Karaletsos, Manuel Wüthrich, Francesco Locatello, Ossama Ahmed, Frederik Träuble, and everyone at MPI for useful discussions.  

\nocite{langley00}

\bibliography{example_paper}
\bibliographystyle{icml2021}

\clearpage
\appendix
\begin{figure*}[t!]
    \centering
    \includegraphics[scale=0.5]{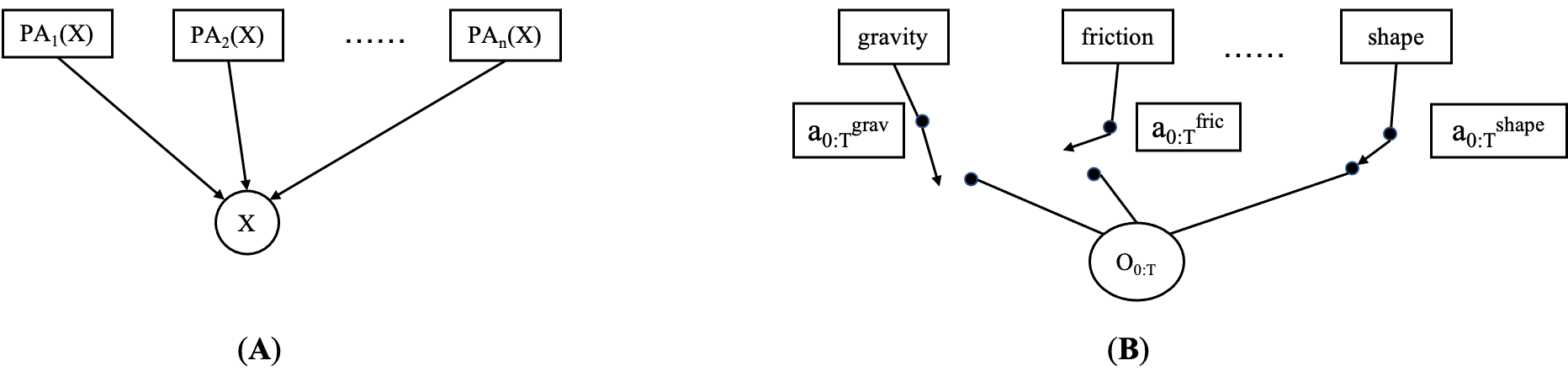}
    \caption{Directed Graphical Model. A Directed Acyclic Graph (DAG) visually represents the causal dependencies of observed and unobserved variables. In \textbf{(A)}, an observed variable $\vb{X}$ is caused by unobserved causal variables, $PA_i(\vb{X})$. In \textbf{(B)}, describes the scenario modeled in the paper, where a subset of the unobserved parent causal variables influence the observed variable $\vb{O}$. The action sequence $\vb{a}_{0:T}$ serves a gating mechanism, allowing or blocking particular edges of the causal graph.}
    \label{fig:CausalGraph}
\end{figure*}
\section{Implementation Details for Experiment Discovery}
\label{TrainingDeets}
\subsection{Planner}
The Experiment Planner consisted of a uniform distribution planner for a horizon of 6 control signals. The planner was trained using the Cross Entropy Method Model Predictive Control \cite{camacho2013model, de2005tutorial} on the true environment. We sampled 30 plans per iteration from the distribution initialized to uniform $\mathcal{U}(controlLow, controlHigh)$. Each of the sampled plans are applied to each of the training environments and the top 10\% of the plans are used to update the distribution. The CEM training required 10 iterations.  
\begin{figure*}[t]
    \centering
    \includegraphics[scale=0.5]{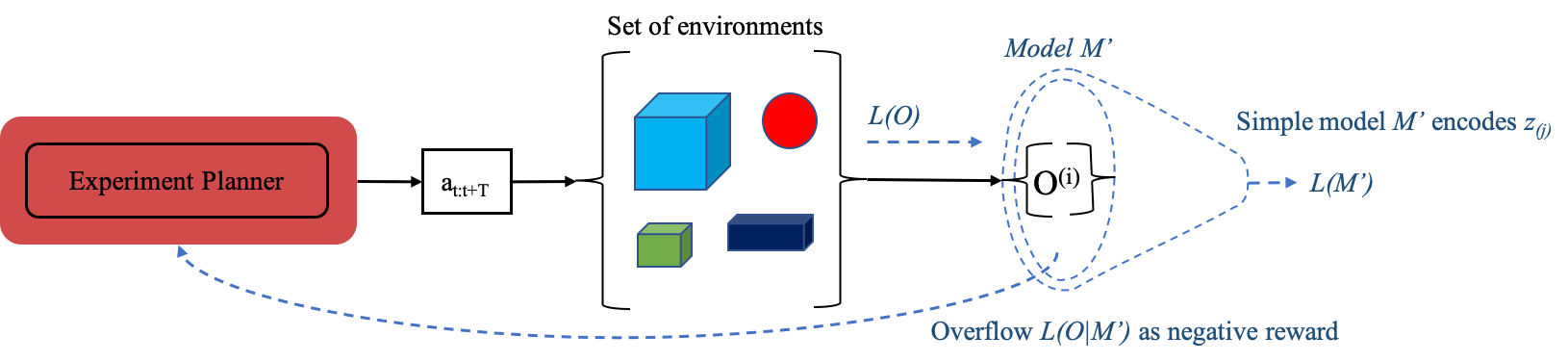}
    \caption{Overview of training. The experiment planner generates a trajectory of actions which is applied to each of the environments with varying causal factors namely mass, shape and size of blocks. For each environment, an observation trajectory or state $\vb{O}^{i} \in O$ is obtained. A simple model with fixed low expressive power is used to approximate the generative model for $\vb{O}$. The "information overflow" $ L(\vb{O}|\vb{M})$ is returned as negative reward forcing $O$ to be caused by few causal factors. }
    \label{fig:trainingLoop}
\end{figure*}

\subsection{Training Environments}
\label{subsec:TrainEnv}
The training environments vary in each experiment. In Section \ref{4.3}, we utilize 3 setups, \texttt{Mass}, \texttt{SizeMass} and \texttt{ShapeSizeMass}. For \texttt{Mass}, we allow the agent to access 5 environments with masses varying from 0.1 kg to 0.5 kg. In \texttt{SizeMass}, the agent has access to 30 environments with masses varying uniformly from 0.1 to 0.5 kg and sizes from 0.05 to 0.1 meters. Finally, in \texttt{ShapeSizeMass}, the agent has access to 60 environments, with masses varying uniformly from 0.1 to 0.5 kg, sizes from 0.05 to 0.1 meters and shapes either being cubes or spheres. During experiment discovery, in each environment, the agent has access to the position of the block in the environment along with its quaternion orientation. 

The total number of causal causal factors of each environment are rather large in number due to the fact that the simulator is a complex realistic physics engine. Examples of the causal factors in the environment include gravity, friction coefficients between all on interacting surfaces, shapes, sizes and masses of blocks, control signal frequencies of the environment. However, we only vary 1 during \texttt{Mass}, 2 during \texttt{SizeMass} and 3 during \texttt{ShapeSizeMass}.

\subsection{Curiosity Reward Calculation}
We predetermine the minimum description length of the clustering model $L(\vb{M})$ by assuming that the observations $\vb{o}_{0:T}$, obtained by applying experimental behavior $\vb{a}_{0:T}$ are produced by a bi-modal generator distribution, where each mode corresponds to either a low or high (quantized) value of a causal factor. This also ensures that $L(\vb{M})$ is as small as possible. 
The planner, eq.~(\ref{eq:new_mdl}) solves the following optimization problem:
\begin{equation}
\label{eq:implementation_distance_cost}
\begin{split}
  \argmax_{\vb{a}_{0:T} \in \mathcal{A}^T}[\min \{d(\vb{o}_{0:T}, \vb{o'}_{0:T}): \vb{o}_{0:T}\in C_1, \vb{o'}_{0:T}\in C_2\} - \\ \max\{d(\vb{o}_{0:T}, \vb{o''}_{0:T}): \vb{o''}_{0:T}, \vb{o}_{0:T}\in C_1\} - \\
  \max\{d(\vb{o'}_{0:T}, \vb{o'''}_{0:T}): \vb{o'}_{0:T}, \vb{o'''}_{0:T}\in C_2\}] \end{split}
\end{equation}
the distance function $d(\cdot, \cdot)$ in the space of trajectories is set to be Soft Dynamic Time Warping \cite{cuturi2017soft}. The trajectory length $T$ is 6 control steps long. The objective is a modified version of the Silhouette Score \cite{rousseeuw1987silhouettes}.

Intuitively, Objective (\ref{eq:implementation_distance_cost}) expresses the ability of a low complexity model, assumed to be bi-modal, to encode the state $\vb{O} = \vb{o}_{0:T}$. If multiple causal factors control $\vb{O}$, then the Minimum Description Length of $L(\vb{O})$ will be high. Subsequently, since $\vb{M}$ is a simple model, the deviation of $\vb{O}$ from $\vb{M}$ will be high i.e. $L(\vb{O}|\vb{M})$ will be high resulting in a low value of the optimization objective.
$C_1$ and $C_2$ correspond to clusters of outcomes which quantize the values of a causal factor isolated by $\vb{a}_{0:T}$. $\vb{o}_{0:T},\vb{o''}_{0:T} \in C_1$ correspond to trajectories of states i.e. observations obtained by applying $\vb{a}_{0:T}$ to environments with say, low values of a causal factor while $\vb{o'}_{0:T},\vb{o'''}_{0:T} \in C_2$ correspond to trajectories of observations i.e. state obtained by applying $\vb{a}_{0:T}$ to environments with say, high values of the same causal factor. Objective (\ref{eq:implementation_distance_cost}) attempts to ensure that these clusters are far apart from each other and are tight i.e. a simple model $\vb{M}$ encodes $\vb{O}$ well.

\section{Implementation Details for Transfer}
\label{TranDeets}
In Section \ref{4.4}, we show the utility of learning causal representations in 2 separate experimental setups. During \texttt{TransferMass}, the agent has access to 10 environments during training, with masses ranging from 0.1 to 0.5 kg. At test time, the agent is required to perform the place-and-orient task masses 2 masses - 0.7 kg and 0.75 kg.
During \texttt{TransferSizeMass}, the agent has access to 10 environments during training, with sizes from either 0.01 or 0.05 m and masses ranging from 0.1 to 0.5 kg. At test time the agent is asked to perform the task on 2 environments with masses 0.7 kg and 0.75 kg with sizes = 0.05 m. 

We find that testing with large and light blocks increase the chances of accidental goal completions. Thus, during test-time, we use environments with high masses for out-of-distribution testing. The causal representation is concatenated to the state of the environment as a contextual input and supplied to a PPO-Optimized Actor-Critic Policy i.e., it receives 57 dimensional input for \texttt{TransferMass}, and a 58 dimensional for \texttt{TransferSizeMass}). The policy network consists of 2 hidden layers with 256 and 128 units respectively. The experiments  are parallelized on 10 CPUs and implemented using stable baselines \cite{stable-baselines}. 
The PPO configuration was \{"gamma":0.9995, "n\_steps": 5000, "ent\_coef": 0, "learning\_rate": 0.00025, "vf\_coef": 0.5, "max\_grad\_norm": 10, "nminibatches": 1000, "noptepochs": 4\}

The agent receives a dense reward at each time step during the maximizing external reward phase (Figure \ref{fig:InferenceLoop}), the negative of the distance of the block from the goal position scaled by factor of 1000. The control signal was repeated 10 times to the actuators of the motors on each finger.

\section{Implementation Details for Pre-trained Behaviors}
\label{sec:downstreamDetails}
In section \ref{downstream}, we study how the acquired experimental behaviors obtained through Causal Curiosity can be used as pre-training for a variety of downstream tasks. The Vanilla CEM depicts the cost of training an experiment planner from scratch to maximize an external dense reward where the agent minimizes the distance between the position of a block in an environment from the goal in the Lifting setup and imparts a velocity to the block along a particular direction in the Travel setup.
\begin{equation}
\label{eq:supervised}
R(\vb{a}_{0:T}) = -\sum_{t}dist(\vb{goal}_t - \vb{block}_t)
\end{equation}
The second baseline (Additive Reward) studies the setup when the agent receives both the curiosity signal and the external reward and attempts to maximize both. The agent receives access all the training environments with varying causal factors and must simultaneously maximize both curiosity and the task reward. The equation below shows the reward maximized for the \texttt{Lifting} task. 
\begin{equation}
\label{eq:Additive}
\begin{split}
R(\vb{a}_{0:T}) = \sum_{envs}\sum^{T}_{t}-dist(\vb{goal}_t - \vb{block}_t)~+ \\ [\min \{d(\vb{o}_{0:T}, \vb{o'}_{0:T}): \vb{o}_{0:T}\in C_1, \vb{o'}_{0:T}\in C_2\} - \\ \max\{d(\vb{o}_{0:T}, \vb{o''}_{0:T}): \vb{o''}_{0:T}, \vb{o}_{0:T}\in C_1\} - \\
  \max\{d(\vb{o'}_{0:T}, \vb{o'''}_{0:T}): \vb{o'}_{0:T}, \vb{o'''}_{0:T}\in C_2\}] 
  \end{split}
\end{equation}
The curious agent first acquired the experimental behavior by interacting with multiple environments with varying causal factors. The lifting skill was obtained during \texttt{Mass}, when the agent attempted to differentiate between multiple blocks of varying mass. The curious agent trained for $600,000$ time steps on the curiosity reward. The acquired behavior was then applied to the downstream lifting task and fine tuned to external rewards. The Vanilla CEM baseline had an identical structure to that of the Curious agent, and received only external reward as in Equation (\ref{eq:supervised}). The additive agent simultaneously optimized both external reward and the curiosity reward as in Equation (\ref{eq:Additive}).

We find that maximizing the curiosity reward in addition to simultaneously maximizing external rewards results in suboptimal performance due to our formulation of the curiosity reward. To maximize curiosity, the agent must discover behaviors that divide environments into 2 clusters. Thus in the context of the experimental setups, this corresponds to acquiring a lifting/pushing behavior that allows the agent to lift/impart horizontal velocity to blocks in half of the environments, while not being able to do so in the remaining environments. However, the explicit external reward incentivizes the agent to lift/impart horizontal velocity blocks in all environments. Thus these competing objectives result in sub-par performance.
\section{Intuition for Definition of Causal Factors}
\label{Gentle}
We begin with a simple example of a person walking on earth. This person experiences various physical processes while interacting in their world, for example gravity, friction, wind etc. These physical processes affect the outcome of interactions of the person with their environment. For example, while jumping on earth, the human experiences gravity which affects the outcome of their jump, the fact that they falls back to the ground. Additionally, these physical processes (or causal mechanisms) are parameterized by causal factors, for example, acceleration constant due to gravity $g = 9.8m/s^2$ on earth, or coefficients of friction between their feet and the ground which assume particular numerical values. 

These causal factors may vary across multiple environments. For example, the person may walk on sand or on ice, surfaces with varying frictional values. Thus the outcome of running on such surfaces will vary, running on sand will require significant effort, while running on ice may result in the person slipping. Thus the coefficient of friction between the person’s feet and the surface they walk on affects the outcome of a particular behavior in said environment. In our definition, $\vb{h}_j$ are causal factors such as friction or gravity etc. $H$ is the global set containing all such causal factors. 

Now we ask the question (which we subsequently answer), given multiple environments, how would a human characterize each of them depending on the value of a causal factor? Through experimental behaviors. The human in the above example would attempt to run in each of the environments she encountered, be it on sand, on ice, in mud etc. If they slipped in an environment, she would characterize it as slippery. If they didn’t, they would characterize it as non-slippery. We attempt to equip our agent with similar logic. The “sequence of actions” ($\vb{a}_{0:T}$) described in our paper corresponds to the human running. The sequence of observations ($\vb{o}_{0:T}$) corresponds to the outcome of running "experiment". $\vb{o}_{0:T}$ might belong to either of the clusters of outcomes $C_a$ or $C_b$ corresponding to slipping or not slipping. 


\begin{algorithm}[tb]
\caption{Inference Loop}\label{alg:inference}
\begin{algorithmic}
\STATE Input: Unseen Test Environment env, trained Planner and Causal Inference Module 
\STATE Initialize $\vb{causalRep} = [~]$
\STATE Initialize training environment set $Envs$
\FOR{k in range(K)}
\STATE Reset env
\STATE Sample experimental behavior $\vb{a}_{0:T}$ $\sim$ CEM($\cdot$|~$\vb{causalRep}$)
\STATE Apply $\vb{a}_{0:T}$ to env
\STATE Collect $\vb{O} = \vb{o}_{0:T}$
\STATE Use learnt $q_M(\vb{z}|\vb{O},\vb{causalRep})$ for cluster assignment i.e. $\vb{z}_k = q_M(\vb{z}|\vb{O}, causalRep)$
\STATE Append $z_k$ to $causalRep$

\ENDFOR
\STATE Learn a policy conditioned on causal factors $\vb{a}_t \sim \pi(\cdot | \vb{o}_t, \vb{z}_{0:K})$ to maximize external reward. 

\end{algorithmic}
\end{algorithm}

\section{Scalability Limitation}
\label{sec:ofat}
We utilize the extremely popular One-Factor-at-a-time (OFAT) general paradigm of scientific investigation, as an inspiration for our method. In the case of many hundreds of causal factors, the complexity of this method will scale exponentially. However, we believe that this would indeed be the case given a human experimenter attempting to discover the causation in any system she is studying.  Learning about causation is a computationally expensive affair. We point the reader towards a wealth of material on the design of scientific experiments and more specifically the lack of scalability of OFAT \cite{fisher1936design, hicks1964fundamental, czitrom1999one}. Nevertheless, OFAT remains the de facto standard for scientific investigation. 

\end{document}